\documentclass[letterpaper, 10 pt, conference]{ieeeconf}
\IEEEoverridecommandlockouts

\usepackage{amsmath,amssymb,amsfonts}
\usepackage{algorithmic}
\usepackage{booktabs}
\usepackage{graphicx}
\usepackage[nolist]{acronym}
\usepackage{algorithm}
\usepackage{algorithmic}
\usepackage{url}
\title{\LARGE \bf
The Radiance of Neural Fields: Democratizing Photorealistic and Dynamic Robotic Simulation}

\author{Georgina Nuthall$^{1}$, Richard Bowden$^{2}$ and Oscar Mendez$^{3}$%
\thanks{$^{1}$ Georgina Nuthall is with the Center of Vision, Speech and Signal Processing, University of Surrey, Guildford, UK. {\tt\small galcoladonuthall@surrey.ac.uk}}%
\thanks{$^{2}$ Richard Bowden is with the Center of Vision, Speech and Signal Processing, University of Surrey, Guildford, UK. {\tt\small r.bowden@surrey.ac.uk}}%
\thanks{$^{3}$ Oscar Mendez is with Locus Robotics, Wilmington, Massachusetts 
USA. {\tt\small omendez@locusrobotics.com}}%
}

\begin{document}
\maketitle
\begin{figure*}[t]
    \centering
    \includegraphics[width=\textwidth]{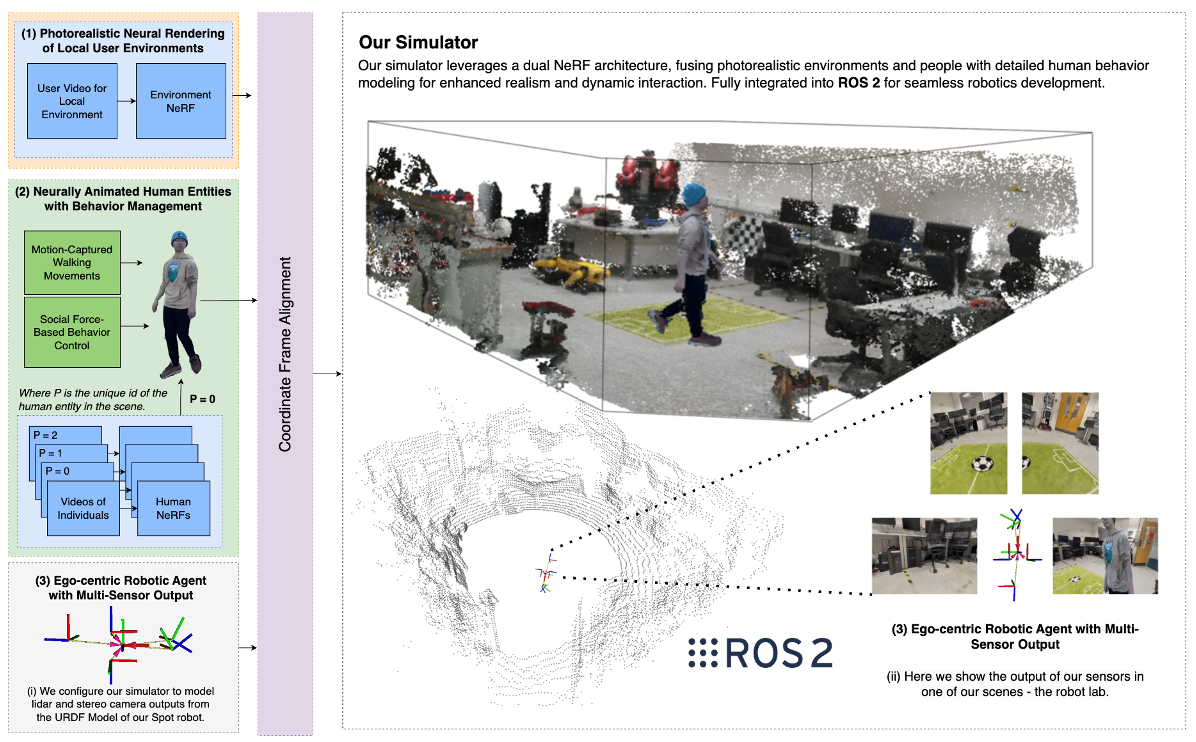} 
    \vspace{-2em} 
    \caption{\label{fig:main}System Overview – The figure illustrates the components of our simulation pipeline. (1) Shows the training of a local environment based on user-provided video footage. (2) Depicts the integration of human entity representations, trained using motion capture data and the social force model, for dynamic human behavior simulation. (3) (i) Details the integration of a predefined robot, exemplified here by Spot’s URDF, into the simulator. All elements are aligned within the same coordinate frame to enable accurate multi-sensor output rendering (3) (ii). The bottom left shows simulated LiDAR output, while the bottom right presents the simulated RGB stereo output.}
\end{figure*}
\thispagestyle{empty}
\pagestyle{empty}
\begin{acronym}[RANSAC]
    \acro{cnn}[CNN]{Convolutional Neural Network}
    \acro{bn}[BN]{Batch Normalisation}
    \acro{rnn}[RNN]{Recurrent Neural Networks}
    \acro{sfm}[SfM]{Structure from Motion}
    \acro{nerf}[NeRF]{Neural Radiance Field}
    \acro{sofm}[SoFM]{Social Force Model}

    \acro{psnr}[PSNR]{Peak Signal-to-Noise Ratio}
    \acro{ssim}[SSIM]{Structural Similarity Index}
    \acro{lpips}[LPIPS]{Learned Perceptual Image Patch Similarity}
\end{acronym}

\begin{abstract}
As robots increasingly coexist with humans, they must navigate complex, dynamic environments rich in visual information and implicit social dynamics, like when to yield or move through crowds. Addressing these challenges requires significant advances in vision-based sensing and a deeper understanding of socio-dynamic factors, particularly in tasks like navigation. To facilitate this, robotics researchers need advanced simulation platforms offering dynamic, photorealistic environments with realistic actors. Unfortunately, most existing simulators fall short, prioritizing geometric accuracy over visual fidelity, and employing unrealistic agents with fixed trajectories and low-quality visuals. To overcome these limitations, we developed a simulator that incorporates three essential elements: (1) photorealistic neural rendering of environments, (2) neurally animated human entities with behavior management, and (3) an ego-centric robotic agent providing multi-sensor output. By utilizing advanced neural rendering techniques in a dual-NeRF simulator, our system produces high-fidelity, photorealistic renderings of both environments and human entities. Additionally, it integrates a state-of-the-art \ac{sofm} to model dynamic human-human and human-robot interactions, creating the first photorealistic and accessible human-robot simulation system powered by neural rendering. The code for the simulator is available at \url{https://gitlab.surrey.ac.uk/gn00217/radiance-of-neural-fields-simulator/}.

\end{abstract}

\section{INTRODUCTION}
Robotic simulators have long been indispensable tools for design and testing, allowing researchers and engineers to safely experiment with robotic tasks without risking harm to hardware or the environment. This capability is vital across many domains, but it becomes especially crucial for human-centered tasks where safety and reliability are paramount. Despite their importance, however, existing simulators face significant limitations: they are often costly, lack realism, and fail to account for human factors. These shortcomings hinder the development of accurate and safe testing platforms, particularly in scenarios involving human-robot interaction. 

Modelling simulation environments that closely replicate physical-world settings is both challenging and time-consuming. Traditionally, these environments have been manually created by computer-aided design specialists, a process that demands significant effort. Alternatively, 3D reconstruction techniques, which generate models from 2D images, can be used, but these often result in noisy and incomplete models that lack the fidelity required for realistic simulations. This is especially problematic when testing computer vision approaches to robotics. Widely-used simulators like Gazebo \cite{Gaz}, CARLA \cite{Dosovitskiy2017}, and NVIDIA Isaac \cite{Isaac} either limit users to predefined environments or rely on custom meshes, further restricting flexibility. Moreover, these simulators often neglect photorealism and realistic modeling of human interactions—between people or robots—limiting their effectiveness in conducting reliable research.

To overcome these challenges, we introduce a novel simulation platform designed to democratize advanced robotic research. This platform models complex dynamics in photorealistic environments with integrated multi-sensor capabilities, enabling accurate testing of robotic tasks, including those in populated environments. By emphasizing realism, accessibility, and ease of use, it reduces costs to developing full-scale digital twins and supports cutting-edge robotics research, particularly in computer vision. Our platform features three key features: (1) Photorealistic Neural Rendering using \acp{nerf} to create immersive, high-fidelity environments; (2) Animated Human Entities with Behavior Management for lifelike human-robot interactions; and (3) An Ego-Centric Robotic Agent with Multi-Sensor Integration that can perform tasks like Simultaneous Localization and Mapping. These features enable researchers to generate realistic, localized environments with accurate sensor data and human behaviors for comprehensive testing.

Recent advancements in 3D scene representation, especially with neural rendering, have gained significant attention. \acp{nerf} are capable of producing photorealistic scenes while implicitly learning key metrics such as depth and occupancy. Although some studies have explored the use of \acp{nerf} in simulation \cite{Ge2022, Byravan2022}, their applications have been limited. To our knowledge, we are the first to present a complete human-robot simulator powered by neural rendering. In this work, we integrate neural rendering into a simulation pipeline to offer a novel, efficient approach to simulating localized environments for robotics research. The main contributions of our paper are the following: 

\begin{enumerate} 

	\item We experiment with low computational cost methods to obtain high-fidelity representations of indoor scenes using the latest approaches in Neural Radiance Fields.
	      
	\item We provide a method for photorealistic rendering of human agents, along with customisable behaviour and reactions to the robotic agents in the scene.
	      
	\item We provide an indoor dataset of multiple sequences captured by a Boston Dynamics advanced quadruped Spot; which we used to benchmark our simulator.
	      
\end{enumerate}

\section{RELATED WORK}
\subsection{Robotic Simulators} 
Robotic simulators are essential for designing and testing robots safely and cost-effectively in a controlled environment. Simulators have been developed for various applications, such as mobile robotics and manipulation n \cite{Koenig2004} \cite{Michel2004} \cite{Rohmer2013} \cite{Todorov2012}, but realism remains a significant challenge \cite{Afzal2020}. For example, AIRSIM \cite{Shah2018} and CARLA \cite{pmlr-v78-dosovitskiy17a} offer photorealistic rendering using Unreal Engine but are limited to autonomous driving and handcrafted environments \cite{Collins2021}. Multi-purpose simulators like GAZEBO \cite{Koenig2004} also struggle with realism and complex environment setup \cite{Afzal2020}, as creating 3D assets manually is time-consuming and error-prone. These errors affect both environmental representation and simulated sensor outputs \cite{OpenGL} \cite{Parker2010} \cite{Gschwandtner2011}. Without realistic rendering of the physical test environment, roboticists are often left with the choice between using synthetic and approximate environments or investing in the costly process of digital twinning \cite{Isaac}. In our work, we address this challenge by focusing on developing an accessible, high-fidelity simulation platform for ego-centric robotic mobile agents.

\subsection{Neural Rendering Approaches} 

Recent advancements in 3D scene representation, particularly through neural rendering techniques, have attracted significant attention due to their photorealistic capabilities. This interest was initially driven by the development of implicit representations using \acp{nerf} \cite{Mildenhall2021}, followed by the introduction of explicit representations through Gaussian Splatting \cite{kerbl3Dgaussians}. While Gaussian Splatting offers faster performance, parametric models such as \acp{nerf} exhibit superior generalization to unseen views \cite{he2024nerfsgaussiansplats}. This attribute is crucial for enhancing the usability of our simulator, which supports the creation of localized environments from handheld video footage captured by users. However, we maintain that our simulator is agnostic to neural rendering approaches.

Early implementations of \acp{nerf} boasted high fidelity results but were costly in both their need for computational resources and training time \cite{Mildenhall2021}. These costs have been reduced significantly through multi-resolution hash encoding \cite{Muller} permitting models to be trained in seconds using only a single GPU. In our dual-\ac{nerf} simulator, we train custom simulation environments on our background \ac{nerf} that uses this approach to speed up the waiting time needed to reconstruct the simulated environment.  

Research on \acp{nerf} has covered a wide range of topics, including few-shot training \cite{Chen2021, Jain2021, Deng2022}, editing \cite{Yuan2022, Wang2022}, and semantics \cite{Zhi2021, Vora2021}. More recently \acp{nerf} have been extended to represent scenes but also humans. ST-NeRF \cite{zhang2021stnerf} is able to render dynamic humans and change their position based on the time sequence video it was trained on. However this means it requires videos of specific actions in order to pose the human models. NeuMan \cite{neural-human-radiance-field}, on the other hand, is able to render unseen human poses by training the \ac{nerf} using a body skeleton and is therefore able to leverage motion capture data to dynamically change the position of a person. Building on these advancements, we integrate this sophisticated human modeling technique into our simulator.

\subsection{Robot Navigation in Crowds} 

There is a growing need in robotics research to develop systems that work seamlessly with humans, especially in assistive and industrial settings. To meet this demand, it is essential to create detailed simulation environments that accurately represent both human behavior and the interactions of people around robotic agents. Such simulations are crucial for ensuring that robots can navigate and operate safely in crowded environments, thereby reducing the risk of human injury and improving the effectiveness of collaborative robotics. It is therefore important to simulate not only photorealistic humans but also the realistic behaviors. PedSimROS\footnote{https://github.com/srl-freiburg/pedsim\_ros} uses the \ac{sofm} \cite{Helbing1998} to determine crowd movement and use Gazebo and RVIZ to render humans a simple non-realistic markers. MengeROS\footnote{https://github.com/ml-lab-cuny/menge\_ros} \cite{Aroor2018}, although provides no advanced rendering for the crowd simulation, does have more advanced features by allowing to set different collision-avoidance strategies like Optimal Reciprocal Collision Avoidance (ORCA) and Pedestrian Velocity Obstacle (PedVO) \cite{ORCA} \cite{Curtis2014}. Crowdbot\footnote{https://crowdbot.eu/CrowdBot-challenge/} and SEAN \footnote{https://sean.interactive-machines.com/} both boast photorealistic human rendering as they make use of the Unity game engine \cite{Grzeskowiak2021} \cite{Tsoi2022}. Inspired by advancements in \ac{nerf}-based methods, which can render photorealistic humans in various poses once trained on specific individuals \cite{neural-human-radiance-field} \cite{weng_humannerf_2022_cvpr}, and the capabilities of HuNavSim\footnote{https://github.com/robotics-upo/hunav\_sim} \cite{Hu}, which uses the SOFM model to simulate group behaviors and dynamic reactions to robot interactions, we leverage these approaches to enhance our simulator. While HuNavSim effectively models group dynamics and individual behaviors based on emotional responses, it lacks photorealistic human renderings. To address this, we applied photorealistic rendering techniques and utilized a skeleton-conditioned NeRF \cite{neural-human-radiance-field} to achieve greater flexibility and realism in human modeling. By integrating these capabilities in addition to photorealistic neural rendering of environments and an ego-centric robotic agent providing multi-sensor output, we are able to create the first photorealistic and accessible human-robot simulation system powered by neural rendering. This system offers a high level of realism and interactivity that is crucial for effective testing and development.

\section{METHODOLOGY}
Central to our simulator is the use of a dual-\ac{nerf} photorealistic rendering framework which provides lifelike, localized environments with dynamic human entities. It incorporates neurally animated human entities and manages a wide range of human actions and interactions for adaptive responses. An ego-centric robotic agent with optional sensors provides detailed multi-sensor outputs for accurate testing of interactions with humans and the environment. Together, these elements form a robust, cost-effective platform that balances high realism with accessibility, supporting advanced research in robotics.

\subsection{Neural Rendering of Local Environments} 

A primary objective of this work is to create a photorealistic simulator that is accessible to the broader robotics community. To achieve this, we utilize a state-of-the-art \ac{nerf} framework \cite{Tancik2023} for modeling local environments. This framework leverages handheld video data captured by users and metrically scaled camera poses, which can be easily acquired using readily available tools such as PolyCam\footnote{https://poly.cam/}.

\subsubsection{Neural Radiance Fields}  

Recent years have seen significant attention given to methods capable of achieving photorealistic view synthesis. \ac{nerf} models use a 5D function using neural networks. The five dimensions are a 3D point in space and a 2D viewing direction often represented as the following.

\begin{equation} 
F(x,\theta, \phi)\xrightarrow{}(\textbf{c}, \sigma)  
\end{equation}  

where $x \in \mathbb{R}^3$ is the point in space and $\theta, \phi$ represent the azimuthal and polar viewing angles. The radiance field describes the color $\mathbf{c}$ represented as $(r, g, b)$ and volume density $\sigma$ for every point and viewing direction of a captured scene. It uses ReLU activations and consists of two branches. One branch is used to estimate the volume density independent of the viewing direction, as the line of sight towards an object should not impact its occupancy. Meanwhile, the other branch of the network is used to estimate the color dependent on the viewing direction and the 3D point in space, which supports viewing specularities within a scene.  

\subsubsection{Our \ac{nerf}}  

The model utilizes techniques that reduce training times and memory requirements \cite{Muller}, ensuring that our simulator enables quick and efficient environment setup. Furthermore, our method uses a  proposal network sampler \cite{Barron2021New} concentrating samples that provide the most value to the final representation, providing higher quality results. The background model, based on NeRFacto \cite{Tancik2023}, also leverages features from NeRF-- \cite{Wang2021}, NeRF-W \cite{MB} and Ref-NeRF \cite{Verbin2021}. We added further inference approaches to speed up the sensor renderings from our model to support real-time capable robotic simulation (see SECTION \ref{sec:system}). 

\subsection{Animated Human Entities with Behavior Management} 

Our model utilizes two \ac{nerf} networks—one for human entities and one for the surrounding scene—integrated within a behavior management framework controlled by a \ac{sofm} pipeline. The human \ac{nerf} \cite{neural-human-radiance-field} is responsible for encoding the appearance and geometry of individuals in the environment, with their poses dynamically generated by the \ac{sofm}. Meanwhile, the scene \ac{nerf} captures the background and environment details, ensuring photorealistic renderings. The training process is sequential, starting with the scene \ac{nerf} to establish an accurate environmental representation, followed by the human \ac{nerf}, which is conditioned on the previously trained scene.

The positioning of individuals within the environment is dictated by the \ac{sofm} alongside predefined emotional states of the entities \cite{Hu}. Human actions, such as walking, are simulated through cycling skeletal poses derived from motion capture data, including sequences from the AMASS dataset \cite{AMASS:ICCV:2019}. By combining neural rendering with human behavior modeling, our framework offers the ability to simulate realistic human interactions within dynamic, photorealistic environments. This enables advanced robotic task testing in scenarios that closely mimic the complexities of real-world settings. The integration of neural rendering and socio-dynamic modeling provides an essential level of realism, crucial for the development and validation of robotics systems intended to operate in environments rich with human interactions and real-world dynamics.

\subsection{Ego-centric Robot Agent with Multi-Sensor Integration} 
Our simulator models various ego-centric sensors for robots, including Spot’s\footnote{https://bostondynamics.com/products/spot/}, which features both appearance sensors like stereo cameras and spatial sensors such as lidar.

\subsubsection{Appearance Sensors} 
\label{sec:appearance}
From our model representing the 3D background simulation environment, we are able to generate RGB cameras using volume rendering. Given the position of the camera to be rendered within the simulation environment, we can cast rays through each pixel of an (n by m) image from the camera origin \textbf{o}. We use the same sampling method as above and obtain the color and density of each sample point. The color of the ray is the weighted average, represented by a transparency $\alpha_{i}$ of each sample along the ray, the probability that the sample is not impeded i.e. its transmittance $T_{i}$ and the evaluated color for the that sample ${c_{i}}$. For each sampled point and viewing direction a color can be approximated as: 

\begin{equation}
    \hat{C}(\textbf{r})=\sum_{i=1}^{N}\alpha_{i}T_{i}\textbf{c}_{i}
\end{equation}

  where  \begin{equation}
  \alpha_{i} = 1 -\exp(\sigma_{i}\delta{i})
  \end{equation}

and \begin{equation}
    T_{i} = \exp(-\sum_{j=1}^{i-1}\sigma_{j} \delta_{j})
    \end{equation}  

The transparency factor of a sample point is the result of an activation function of its estimated density and distance from itself \textit{i} to the unimpeded next sample \textit{i+1} represented by $\delta_{i}$. Thereby obtaining an RGB value for every pixel in the image allowing the simulator to accurately replicate the output of real-world camera sensors in the robot's perception system.

\subsubsection{Spatial Sensors} 

We offer users the option to output data from multiple depth sensors. Building on the methodology outlined in SECTION \ref{sec:appearance}, an approximate depth can be calculated for a single ray by using the accumulated transmittance of samples, which can then generate depth images. This ray-based depth estimation approach can also be extended to simulate LiDAR. In our simulator, users can choose from various LiDAR configurations based on the number of channels and the vertical field of view (FoV). By default, we model a 16-beam LiDAR with a vertical FoV ranging from -15 to 15 degrees. Leveraging the trained environment model, the simulator can render a point cloud by casting rays in a full 360-degree sweep around the LiDAR, with the rays originating from the LiDAR’s center \textbf{w} and direction \textbf{d} 

\begin{equation} 
    \textbf{d} = (\cos{\theta}\sin{\phi}, \sin{\theta}\sin{\phi}, \cos{\phi})^\intercal 
\end{equation} 

where $\theta$ is the horizontal angle, which sweeps around the LiDAR in a full 360 degree circle and $\phi$ is the vertical angle of the ray which controls the up-and-down direction of the rays. In our case, the 16 beams are spread evenly from -15° to 15°. We cast N rays every 2$\pi$/N  radians to complete a 360 degree loop within the desired number of samples allowing the simulator to generate a detailed and accurate point cloud that replicates the output of physical-world LiDAR sensors in diverse environments.

\subsection{System Architecture and Components}
\label{sec:system}

Our simulator integrates three core components: (1) photorealistic neural rendering for environment visualization, (2) neurally animated human models with behavior management, and (3) an ego-centric robotic agent equipped with multi-sensor outputs. FIGURE \ref{fig:main} presents an overview of the system and its integration into ROS2. The simulation process begins with the user providing video data and camera poses to build a local environment model, which is then integrated into the simulator. Human entities are added based on predefined social force parameters, with our human entity models pre-loaded. Accurate interaction between the environment and human \acp{nerf} is ensured by aligning their coordinate frames. This allows for the ego-centric agent sensors to accurately output the state of dynamic scene. 

Rendering humans from the robot's perspective is unnecessary if they are out of view of the robot's cameras. In such cases, we reduce rendering times by excluding these humans. We do this by performing an efficient bounding box calculation to determine if the human is visible by the robot and subsequently render only the necessary rays for the human (see ALGORITHM \ref{alg:vis-determination}). This approach allows for an optimized simulator.

\begin{algorithm}
\caption{Visibility Determination via Bounding Box}
\begin{algorithmic}
\label{alg:vis-determination}
\STATE \textbf{Input:} Bounding box coordinates in human coordinates $(\text{Min}, \text{Max})$
\STATE \textbf{Input:} Transformation matrix $\mathbf{T}_{robot \to human}$
\STATE \textbf{Input:} Camera intrinsic matrix $\mathbf{K}$
\STATE \textbf{Input:} Camera image dimensions $(W, H)$

\STATE \textbf{For each corner of the bounding box}
\STATE \quad Transform the corner point to robot coordinates:
\[
\mathbf{p}_{robot} = \mathbf{T}_{robot \to human} \cdot \mathbf{p}_{human}
\]
\STATE \quad Project the point to the camera image plane:
\[
\begin{pmatrix}
u \\
v \\
1
\end{pmatrix} = \mathbf{K} \cdot \begin{pmatrix}
X \\
Y \\
Z
\end{pmatrix}
\]
\STATE \quad \textbf{If} $0 \leq u < W$ \textbf{and} $0 \leq v < H$
\STATE \quad \quad Mark the bounding box as visible
\STATE \quad \textbf{Else}
\STATE \quad \quad Skip rendering of this human
\STATE \textbf{End If}
\STATE \textbf{End For}

\end{algorithmic}
\end{algorithm}

\section{EXPERIMENTS AND RESULTS}
To rigorously evaluate the capabilities and performance of our simulator, a series of experiments were conducted focusing on three critical aspects: photorealistic neural rendering of local scenes, neurally animated human entities, and the ego-centric robotic agent. 

\subsection{Evaluation of Neural Rendering of Local Scenes}
\label{sec:local-scenes}
To assess the quality of neural rendering for local scenes, we captured short videos of diverse environments and obtained camera poses using \acp{sfm} \cite{Schonberger2016}. As detailed in TABLE~\ref{tab:local-env}, our approach to neurally rendered environments shows competitive performance across common environments for robotic agents, including a Robot Lab, Kitchen, and Living Room.
\begin{table}[h]
\centering
\resizebox{\columnwidth}{!}{%
\begin{tabular}{lccc}
\hline
Environment & \acs{psnr} $\uparrow$ & \acs{ssim} $\uparrow$ & \acs{lpips} $\downarrow$ \\ \hline
Robot Lab             & 23.1          & 0.78          & 0.26             \\
Common Room           & 22.1          & 0.80          & 0.25             \\
Kitchen               & 20.7          & 0.79          & 0.29             \\
Living Room            & 21.8          & 0.77          & 0.35             \\
Seminar Room          & 21.2          & 0.79          & 0.33             \\ \hline
\end{tabular}%
}
\caption{RGB Image Quality for multiple simulated environments \vspace{-10pt}}
\label{tab:local-env}
\end{table}
\subsection{Evaluation of Neurally Animated Human Entities}
\begin{figure}[!h]
    \centering
    \vspace{-0.25cm}
    \includegraphics[width=0.5\textwidth]{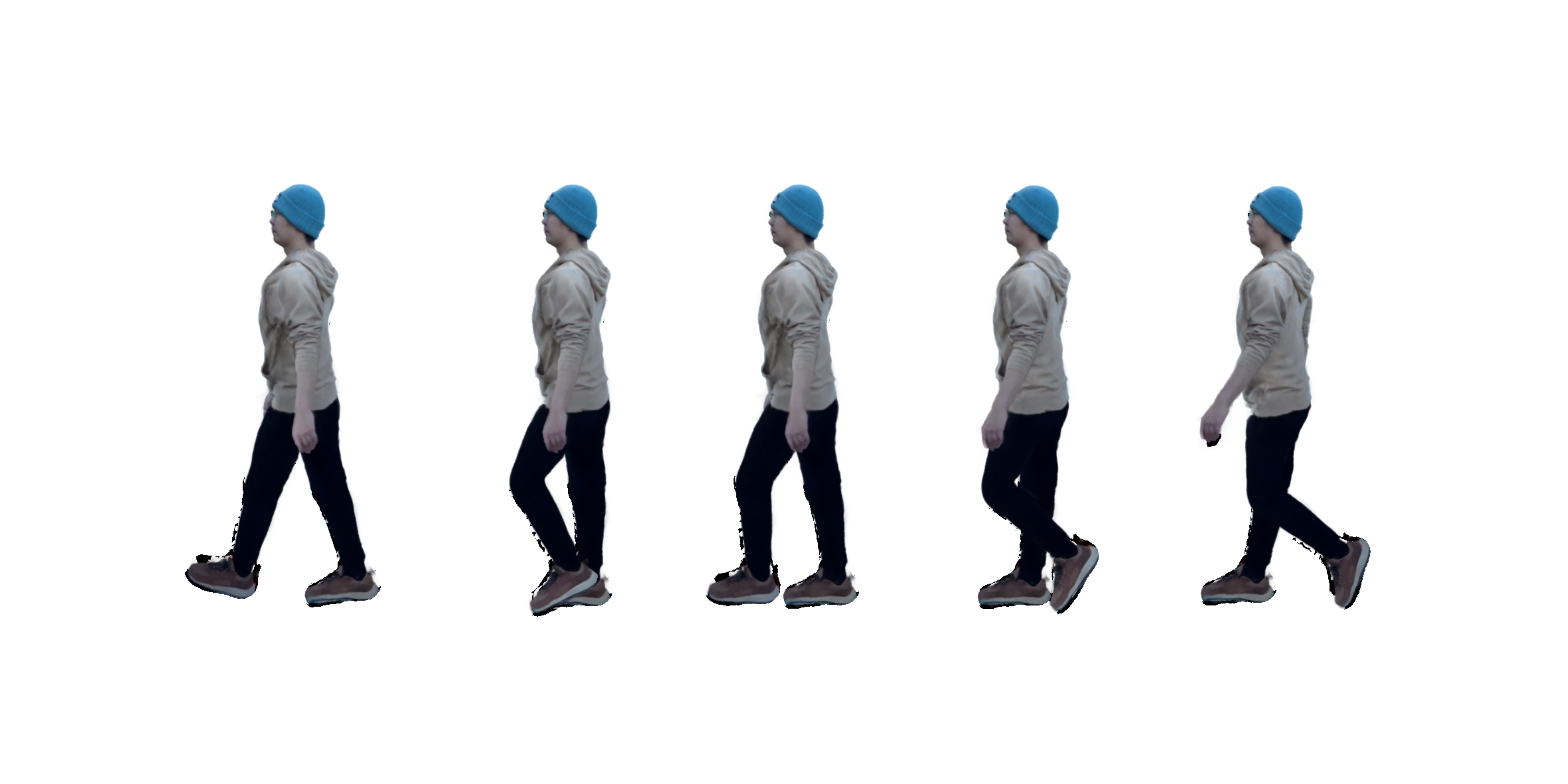}
    \caption{Sample Gait Cycle using Motion Capture \vspace{-15pt}}
    \label{fig:human-nerf}
\end{figure}

FIGURE \ref{fig:human-nerf} shows an example of one of the available photorealistic humans in our simulator.
The agents can be controlled using poses from a motion capture (or similar system). For most of our work, we use a standard gait, shown in FIGURE \ref{fig:human-nerf} which allows the agents to perform realistic motion between the different configured waypoints. However, our work can be articulated with any human skeleton, allowing us to represent the different emotions showcased by the \ac{sofm} model, creating dynamic reactions to robots in the scene.

\subsection{Ego-Centric Robotic Agent Performance}
To evaluate our simulator’s ability to accurately model the sensor outputs of ego-centric robotic agents, we test its performance using the Spot robot by Boston Dynamics. This evaluation focuses on three key aspects: modeling sensor outputs, implementing a vision-based SLAM (Simultaneous Localization and Mapping) method, and performing an object detection task.

\subsubsection{Evaluating Simulator Sensor Output}
To validate our simulator, we collected data from the Boston Dynamics Spot platform, including RGB, Depth, and LiDAR outputs. We captured data from all five stereo cameras for a 360-degree view and used Graph Navigation for initial position estimates, refined with \ac{sfm} techniques. Spot was manually controlled to map the environment and then autonomously followed the recorded trajectory. In the simulation, we retraced these trajectories to compare the appearance and spatial sensor outputs of both the simulated and real Spot.

\begin{figure}[!h]
    \includegraphics[width=0.24\textwidth]{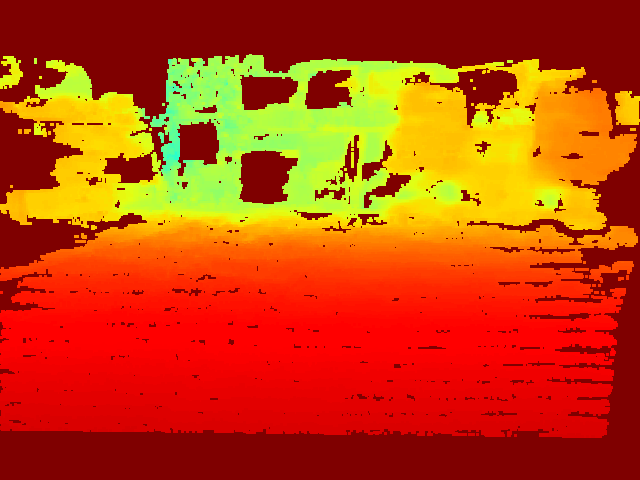}
    \includegraphics[width=0.24\textwidth]{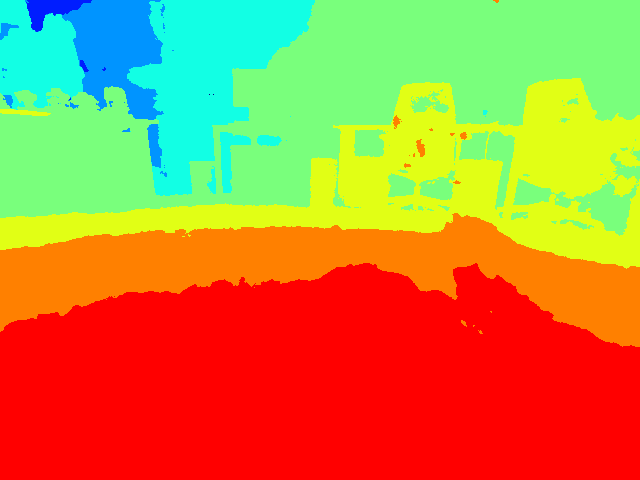}
    \caption{Depth Image Comparison: (Left) Real Back Depth Image and (Right) Simulated Back Depth Image for Comparison}
    \label{fig:spot_depth_renders}
\end{figure}

In order to evaluate the performance of our depth estimation, we follow the evaluation criteria of \cite{spencer2023kick}. TABLE \ref{tab:spot_depth} shows results on the depth cameras mounted on the Boston Dynamics Spot robot. 
FIGURE \ref{fig:spot_depth_renders} shows the qualitative results of the depth renders compared to the ground-truth RGB-D estimates from Spot's depth sensors. 
As can be seen, the data is both accurate and high fidelity, allowing the user to use the depth images as part of a development pipeline. 

\begin{table}[!h]
\centering
\resizebox{\columnwidth}{!}{%
\begin{tabular}{lcccccc}
\hline
Spots Depth Cameras      & AbsRel (\%) & $\delta_{.05}$ & $\delta_{.1}$ & $\delta_{.25}$ & $\delta_{.25^2}$ & $\delta_{.25^3}$ \\ \hline
left fisheye       & 0.25   & 0.26     & 0.29     & 0.58     & 0.86     & 0.93     \\
right fisheye      & 0.24   & 0.258    & 0.29     & 0.581    & 0.88     & 0.95     \\
back fisheye       & 0.27   & 0.24     & 0.28     & 0.56     & 0.84     & 0.92     \\
frontleft fisheye  & 0.6    & 0.215    & 0.23     & 0.3      & 0.698    & 0.74     \\
frontright fisheye & 0.67   & 0.21     & 0.21     & 0.31     & 0.63     & 0.7      \\ \hline
\end{tabular}%
}
\caption{Depth Error for different Spot Robot Simulated Cameras}
\label{tab:spot_depth}
\end{table}

\begin{figure}[!h]
    \includegraphics[angle=270, width=0.24\textwidth]{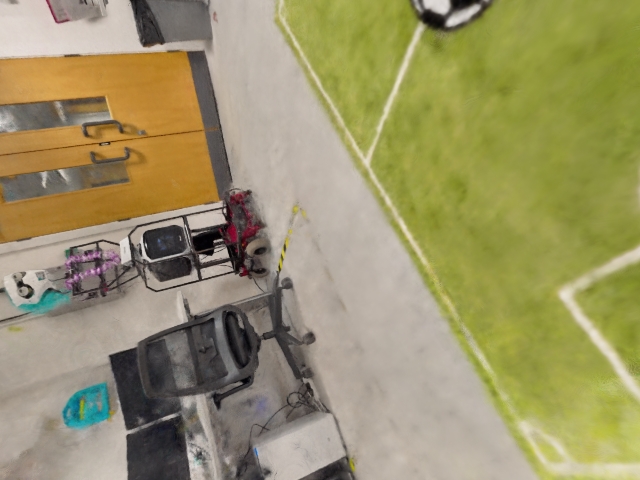}
    \includegraphics[angle=270, width=0.24\textwidth]{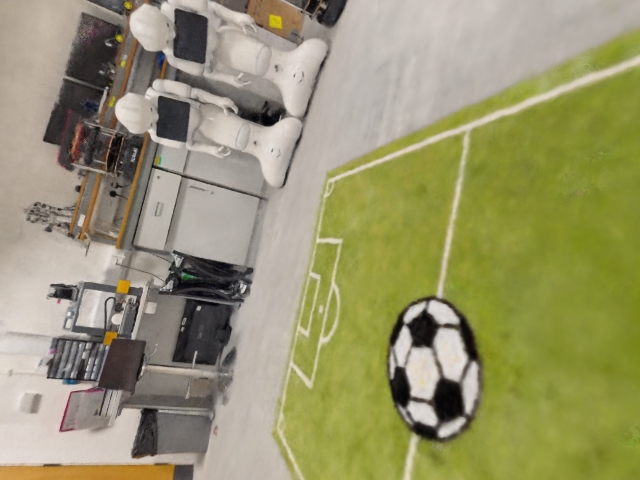}
    \caption{(Left) Front-Left and (Right) Front-Right Simulated Camera Renders}
    \label{fig:cam_renders}
\end{figure}
\begin{table}[!h]
\centering
\resizebox{\columnwidth}{!}{%
\begin{tabular}{lccc}
\hline
Robot Lab RGB Cameras & \acs{psnr} $\uparrow$ & \acs{ssim} $\uparrow$ & \acs{lpips} $\downarrow$ \\ \hline
left fisheye          & 9.5           & 0.28          & 0.55             \\
right fisheye         & 9.37          & 0.28          & 0.59             \\
back fisheye          & 11.8          & 0.49          & 0.42             \\
frontleft fisheye     & 10.1          & 0.34          & 0.53             \\
frontright fisheye    & 10.9          & 0.34          & 0.53             \\ \hline
\end{tabular}%
}
\caption{RGB Image quality for simulated cameras}
\label{tab:spot-sensors}
\end{table}

Furthermore, we evaluate the RGB stereo camera output from our simulated Spot using the same metrics as outlined in SECTION \ref{sec:local-scenes} (see TABLE \ref{tab:spot-sensors}). While some variation in results is expected due to training our background environments with different camera models, this variability does not detract from the overall quality of the simulation. In fact, FIGURE \ref{fig:cam_renders} highlights the impressive level of photorealism achieved within our custom NeRF-based simulation environments. This not only underscores the robustness of our approach but also showcases the versatility of our rendering technique, which consistently produces realistic and visually compelling outputs across varying conditions.

\subsubsection{SLAM Performance Analysis}
The vision-based SLAM method is essential for robots navigating dynamic environments, enabling them to construct and update maps while tracking their own position. This capability is particularly important for robots operating in complex, unstructured spaces, such as those found in assistive applications. 

We ran ORBSLAM3 \cite{Campos2021} on matching trajectories of RGB data from each of Spot's five cameras, both simulated and real. The results showed that the median average trajectory error for Spot's real cameras was 0.14 meters, compared to 0.24 meters in our simulated environment. This indicates a difference of just 0.1 meters between the simulated and real-world testing, highlighting the close alignment between our simulation and actual performance. In further ablations we found that we could enhance the accuracy by incorporating a proportion of the real-world camera data into the training of our simulation background environment, underscoring the potential for continued improvement and refinement of our simulation pipeline.

\subsubsection{Human-Robot Task: Object Detection}
\begin{figure}
    \centering
    \includegraphics[width=\columnwidth]{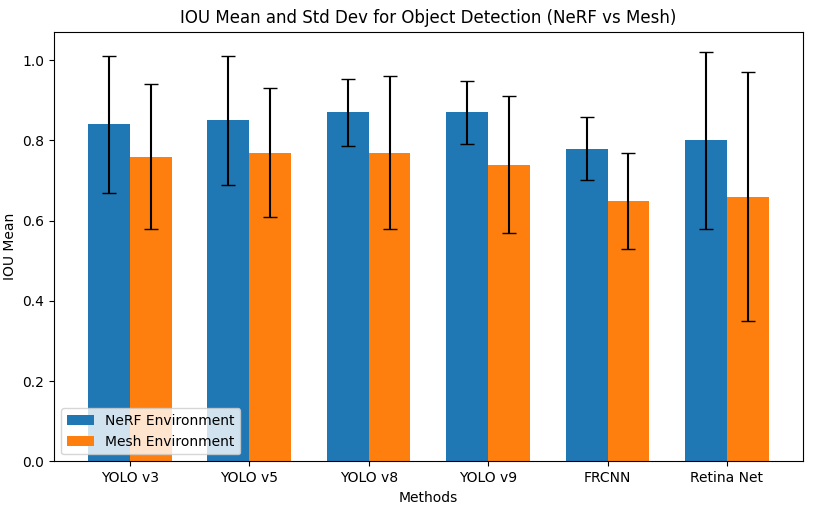}
    \caption{Object Detection Evaluation – Comparison of object detection performance across simulated environments and 3D mesh reconstructions \vspace{-10pt}}
    \label{fig:object-detection}
\end{figure}

Object detection is crucial for assistive robotics, as robots must identify and interact with objects to support users effectively. Evaluating our simulator’s performance ensures it can model real-world challenges, supporting computer vison-based tasks like item retrieval, obstacle avoidance, and daily assistance. To evaluate our simulator's object detection capabilities, we test on 50 diverse indoor environments from the ScanNet++ \cite{yeshwanthliu2023scannetpp} dataset. We compare object detection performance on multiple models \cite{redmon2018yolov3incrementalimprovement, glenn_jocher_2021_5563715, 10533619, wang2024yolov9, girshick2015fastrcnn, lin2018focallossdenseobject} by analyzing RGB images from both the simulated environment and a 3D mesh reconstruction. By comparing bounding boxes from images in both environments, we find that the neurally rendered scenes consistently achieve higher Intersection over Union (IoU) and show less variation across scenes compared to the 3D mesh reconstruction (see FIGURE \ref{fig:object-detection}). This demonstrates that NeRF-based simulations offer superior performance for object detection tasks in robotics. 

\section{CONCLUSIONS}
In this paper, we have demonstrated the capabilities of using Neural Radiance Fields as a robotic simulator that provides multi-sensor readings to validate robotic tasks. High-fidelity simulation with implicitly learnt sensor measurements will be critical for advancing the field of robotics, as it allows researchers and developers to test and refine their algorithms in a safe and controlled environment that is not only representative of the real world, but that allows learning-based vision approaches to be trained in simulation. 

While the current iteration of our simulator provides a solid foundation for robotics research and experimentation, there are several avenues for future development that warrant exploration. One aspect is the integration of a robust physics engine. Incorporating a physics engine will enable our simulator to accurately model real-world interactions between objects, thereby enhancing the fidelity and realism of simulations. 

\bibliography{bibliography}

\begin{thebibliography}{10}

\bibitem{Afzal2020}
Afsoon Afzal, Deborah~S. Katz, Claire~Le Goues, and Christopher~S. Timperley.
\newblock A study on the challenges of using robotics simulators for testing.
\newblock 4 2020.

\bibitem{Aroor2018}
Anoop Aroor, Susan~L. Epstein, and Raj Korpan.
\newblock Mengeros: a crowd simulation tool for autonomous robot navigation.
\newblock {\em AAAI Fall Symposium - Technical Report}, FS-17-01 -
  FS-17-05:123--125, 1 2018.

\bibitem{Barron2021New}
Jonathan~T. Barron, Ben Mildenhall, Dor Verbin, Pratul~P. Srinivasan, and Peter
  Hedman.
\newblock Mip-nerf 360: Unbounded anti-aliased neural radiance fields.
\newblock 2022-June:5460--5469, 11 2021.

\bibitem{ORCA}
Jur Van~Den Berg, Stephen~J. Guy, Ming Lin, and Dinesh Manocha.
\newblock Reciprocal n-body collision avoidance.
\newblock {\em Springer Tracts in Advanced Robotics}, 70:3--19, 2011.

\bibitem{Byravan2022}
Arunkumar Byravan, Jan Humplik, Leonard Hasenclever, Arthur Brussee, Francesco
  Nori, Tuomas Haarnoja, Ben Moran, Steven Bohez, Fereshteh Sadeghi, Bojan
  Vujatovic, and Nicolas~Heess Deepmind.
\newblock Nerf2real: Sim2real transfer of vision-guided bipedal motion skills
  using neural radiance fields.
\newblock 10 2022.

\bibitem{Campos2021}
Carlos Campos, Richard Elvira, Juan~J.Gomez Rodriguez, Jose~M.M. Montiel, and
  Juan~D. Tardos.
\newblock Orb-slam3: An accurate open-source library for visual,
  visual-inertial, and multimap slam.
\newblock {\em IEEE Transactions on Robotics}, 37:1874--1890, 12 2021.

\bibitem{Chen2021}
Anpei Chen, Zexiang Xu, Fuqiang Zhao, Xiaoshuai Zhang, Fanbo Xiang, Jingyi Yu,
  and Hao Su.
\newblock Mvsnerf: Fast generalizable radiance field reconstruction from
  multi-view stereo, 2021.

\bibitem{Collins2021}
Jack Collins, Shelvin Chand, Anthony Vanderkop, and David Howard.
\newblock A review of physics simulators for robotic applications.
\newblock {\em IEEE Access}, 9:51416--51431, 2021.

\bibitem{Curtis2014}
Sean Curtis and Dinesh Manocha.
\newblock Pedestrian simulation using geometric reasoning in velocity space.
\newblock {\em Pedestrian and Evacuation Dynamics 2012}, pages 875--890, 2014.

\bibitem{Deng2022}
Kangle Deng, Andrew Liu, Jun-Yan Zhu, and Deva Ramanan.
\newblock Depth-supervised nerf: Fewer views and faster training for free,
  2022.

\bibitem{Dosovitskiy2017}
Alexey Dosovitskiy, German Ros, Felipe Codevilla, Antonio Lopez, and Vladlen
  Koltun.
\newblock Carla: An open urban driving simulator.
\newblock pages 1--16. PMLR, 10 2017.

\bibitem{pmlr-v78-dosovitskiy17a}
Alexey Dosovitskiy, German Ros, Felipe Codevilla, Antonio Lopez, and Vladlen
  Koltun.
\newblock {CARLA}: {An} open urban driving simulator.
\newblock In Sergey Levine, Vincent Vanhoucke, and Ken Goldberg, editors, {\em
  Proceedings of the 1st Annual Conference on Robot Learning}, volume~78 of
  {\em Proceedings of Machine Learning Research}, pages 1--16. PMLR, 13--15 Nov
  2017.

\bibitem{glenn_jocher_2021_5563715}
Glenn~Jocher et. al.
\newblock {ultralytics/yolov5: v6.0 - YOLOv5n 'Nano' models, Roboflow
  integration, TensorFlow export, OpenCV DNN support}, October 2021.

\bibitem{Ge2022}
Yunhao Ge, Harkirat Behl, Jiashu Xu, Suriya Gunasekar, Neel Joshi, Yale Song,
  Xin Wang, Laurent Itti, and Vibhav Vineet.
\newblock Neural-sim: Learning to generate training data with nerf.
\newblock {\em Lecture Notes in Computer Science (including subseries Lecture
  Notes in Artificial Intelligence and Lecture Notes in Bioinformatics)}, 13683
  LNCS:477--493, 2022.

\bibitem{girshick2015fastrcnn}
Ross Girshick.
\newblock Fast r-cnn, 2015.

\bibitem{Grzeskowiak2021}
Fabien Grzeskowiak, David Gonon, Daniel Dugas, Diego Paez-Granados, Jen~Jen
  Chung, Juan Nieto, Roland Siegwart, Aude Billard, Marie Babel, and Julien
  Pettré.
\newblock Crowd against the machine: A simulation-based benchmark tool to
  evaluate and compare robot capabilities to navigate a human crowd.
\newblock {\em Proceedings - IEEE International Conference on Robotics and
  Automation}, 2021-May:3879--3885, 4 2021.

\bibitem{Gschwandtner2011}
Michael Gschwandtner, Roland Kwitt, Andreas Uhl, and Wolfgang Pree.
\newblock Blensor: Blender sensor simulation toolbox.
\newblock {\em Lecture Notes in Computer Science (including subseries Lecture
  Notes in Artificial Intelligence and Lecture Notes in Bioinformatics)}, 6939
  LNCS:199--208, 2011.

\bibitem{he2024nerfsgaussiansplats}
Siming He, Zach Osman, and Pratik Chaudhari.
\newblock From nerfs to gaussian splats, and back, 2024.

\bibitem{Helbing1998}
Dirk Helbing and Peter Molnar.
\newblock Social force model for pedestrian dynamics.
\newblock {\em Physical Review E}, 51:4282--4286, 5 1998.

\bibitem{Jain2021}
Ajay Jain, Matthew Tancik, and Pieter Abbeel.
\newblock Putting nerf on a diet: Semantically consistent few-shot view
  synthesis, 2021.

\bibitem{zhang2021stnerf}
Zhang Jiakai, Liu Xinhang, Ye~Xinyi, Zhao Fuqiang, Zhang Yanshun, Wu~Minye,
  Zhang Yingliang, Xu~Lan, and Yu~Jingyi.
\newblock Editable free-viewpoint video using a layered neural representation.
\newblock In {\em ACM SIGGRAPH}, 2021.

\bibitem{neural-human-radiance-field}
Wei Jiang, Kwang~Moo Yi, Golnoosh Samei, Oncel Tuzel, and Anurag Ranjan.
\newblock Neuman: Neural human radiance field from a single video, 2022.

\bibitem{kerbl3Dgaussians}
Bernhard Kerbl, Georgios Kopanas, Thomas Leimk{\"u}hler, and George Drettakis.
\newblock 3d gaussian splatting for real-time radiance field rendering.
\newblock {\em ACM Transactions on Graphics}, 42(4), July 2023.

\bibitem{Koenig2004}
Nathan Koenig and Andrew Howard.
\newblock Design and use paradigms for gazebo, an open-source multi-robot
  simulator.
\newblock {\em 2004 IEEE/RSJ International Conference on Intelligent Robots and
  Systems (IROS)}, 3:2149--2154, 2004.

\bibitem{lin2018focallossdenseobject}
Tsung-Yi Lin, Priya Goyal, Ross Girshick, Kaiming He, and Piotr Dollár.
\newblock Focal loss for dense object detection, 2018.

\bibitem{AMASS:ICCV:2019}
Naureen Mahmood, Nima Ghorbani, Nikolaus~F. Troje, Gerard Pons-Moll, and
  Michael~J. Black.
\newblock {AMASS}: Archive of motion capture as surface shapes.
\newblock In {\em International Conference on Computer Vision}, pages
  5442--5451, October 2019.

\bibitem{MB}
Ricardo Martin-Brualla, Noha Radwan, Mehdi~S.M. Sajjadi, Jonathan~T. Barron,
  Alexey Dosovitskiy, and Daniel Duckworth.
\newblock Nerf in the wild: Neural radiance fields for unconstrained photo
  collections.
\newblock {\em Proceedings of the IEEE Computer Society Conference on Computer
  Vision and Pattern Recognition}, pages 7206--7215, 8 2020.

\bibitem{Michel2004}
Olivier Michel.
\newblock Cyberbotics ltd. webots™: Professional mobile robot simulation.
\newblock {\em International Journal of Advanced Robotic Systems}, 1:39--42, 3
  2004.

\bibitem{Mildenhall2021}
Ben Mildenhall, Pratul~P. Srinivasan, Matthew Tancik, Jonathan~T. Barron, Ravi
  Ramamoorthi, and Ren Ng.
\newblock Nerf.
\newblock {\em Communications of the ACM}, 65:99--106, 12 2021.

\bibitem{Muller}
Thomas Müller, Alex Evans, Christoph Schied, and Alexander Keller.
\newblock Instant neural graphics primitives with a multiresolution hash
  encoding.
\newblock {\em ACM Transactions on Graphics}, 41:102, 1 2022.

\bibitem{Isaac}
NVIDIA.
\newblock Isaac platform for robotic.
\newblock
  \url{https://www.nvidia.com/en-gb/deep-learning-ai/industries/robotics/}.
\newblock (accessed: 29.04.2024).

\bibitem{OpenGL}
OpenGL.
\newblock The opengl programming guide.
\newblock \url{http://www.opengl-redbook.com/}.
\newblock (accessed: 29.04.2024).

\bibitem{Parker2010}
Steven~G. Parker, James Bigler, Andreas Dietrich, Heiko Friedrich, Jared
  Hoberock, David Luebke, David McAllister, Morgan McGuire, Keith Morley,
  Austin Robison, and Martin Stich.
\newblock Optix.
\newblock {\em ACM Transactions on Graphics (TOG)}, 29, 7 2010.

\bibitem{Hu}
Noé Pérez-Higueras, Roberto Otero, Fernando Caballero, and Luis Merino.
\newblock Hunavsim: A ros 2 human navigation simulator for benchmarking
  human-aware robot navigation.
\newblock 5 2023.

\bibitem{redmon2018yolov3incrementalimprovement}
Joseph Redmon and Ali Farhadi.
\newblock Yolov3: An incremental improvement, 2018.

\bibitem{Gaz}
Open Robotics.
\newblock Gazebo.
\newblock \url{https://gazebosim.org/home}.
\newblock (accessed: 29.04.2024).

\bibitem{Rohmer2013}
Eric Rohmer, Surya~P.N. Singh, and Marc Freese.
\newblock V-rep: A versatile and scalable robot simulation framework.
\newblock {\em IEEE International Conference on Intelligent Robots and
  Systems}, pages 1321--1326, 2013.

\bibitem{Schonberger2016}
Johannes~L. Schonberger and Jan-Michael Frahm.
\newblock Structure-from-motion revisited.
\newblock In {\em Proceedings of the IEEE Conference on Computer Vision and
  Pattern Recognition (CVPR)}, June 2016.

\bibitem{Shah2018}
Shital Shah, Debadeepta Dey, Chris Lovett, and Ashish Kapoor.
\newblock Airsim: High-fidelity visual and physical simulation for autonomous
  vehicles.
\newblock {\em Springer Proceedings in Advanced Robotics}, 5:621--635, 2018.

\bibitem{spencer2023kick}
Jaime Spencer, Chris Russell, Simon Hadfield, and Richard Bowden.
\newblock Kick back \& relax: Learning to reconstruct the world by watching
  slowtv, 2023.

\bibitem{Tancik2023}
Matthew Tancik, Ethan Weber, Evonne Ng, Ruilong Li, Brent Yi, Justin Kerr,
  Terrance Wang, Alexander Kristoffersen, Jake Austin, Kamyar Salahi, Abhik
  Ahuja, David McAllister, and Angjoo Kanazawa.
\newblock Nerfstudio: A modular framework for neural radiance field
  development.
\newblock 2 2023.

\bibitem{Todorov2012}
Emanuel Todorov, Tom Erez, and Yuval Tassa.
\newblock Mujoco: A physics engine for model-based control.
\newblock {\em IEEE International Conference on Intelligent Robots and
  Systems}, pages 5026--5033, 2012.

\bibitem{Tsoi2022}
Nathan Tsoi, Alec Xiang, Peter Yu, Samuel~S. Sohn, Greg Schwartz, Subashri
  Ramesh, Mohamed Hussein, Anjali~W. Gupta, Mubbasir Kapadia, and Marynel
  Vazquez.
\newblock Sean 2.0: Formalizing and generating social situations for robot
  navigation.
\newblock {\em IEEE Robotics and Automation Letters}, 7:11047--11054, 10 2022.

\bibitem{10533619}
Rejin Varghese and Sambath M.
\newblock Yolov8: A novel object detection algorithm with enhanced performance
  and robustness.
\newblock In {\em 2024 International Conference on Advances in Data Engineering
  and Intelligent Computing Systems (ADICS)}, pages 1--6, 2024.

\bibitem{Verbin2021}
Dor Verbin, Peter Hedman, Ben Mildenhall, Todd Zickler, Jonathan~T. Barron, and
  Pratul~P. Srinivasan.
\newblock Ref-nerf: Structured view-dependent appearance for neural radiance
  fields.
\newblock {\em Proceedings of the IEEE Computer Society Conference on Computer
  Vision and Pattern Recognition}, 2022-June:5481--5490, 12 2021.

\bibitem{Vora2021}
Suhani Vora, Noha Radwan, Klaus Greff, Henning Meyer, Kyle Genova, Mehdi S.~M.
  Sajjadi, Etienne Pot, Andrea Tagliasacchi, and Daniel Duckworth.
\newblock Nesf: Neural semantic fields for generalizable semantic segmentation
  of 3d scenes.
\newblock 11 2021.

\bibitem{Wang2022}
Can Wang, Menglei Chai, Mingming He, Dongdong Chen, and Jing Liao.
\newblock Clip-nerf: Text-and-image driven manipulation of neural radiance
  fields, 2022.

\bibitem{wang2024yolov9}
Chien-Yao Wang and Hong-Yuan~Mark Liao.
\newblock Yolov9: Learning what you want to learn using programmable gradient
  information.
\newblock 2024.

\bibitem{Wang2021}
Zirui Wang, Shangzhe Wu, Weidi Xie, Min Chen, and Victor~Adrian Prisacariu.
\newblock Nerf--: Neural radiance fields without known camera parameters.
\newblock 2 2021.

\bibitem{weng_humannerf_2022_cvpr}
Chung-Yi Weng, Brian Curless, Pratul~P. Srinivasan, Jonathan~T. Barron, and Ira
  Kemelmacher-Shlizerman.
\newblock Human{N}e{RF}: Free-viewpoint rendering of moving people from
  monocular video.
\newblock In {\em Proceedings of the IEEE/CVF Conference on Computer Vision and
  Pattern Recognition (CVPR)}, pages 16210--16220, June 2022.

\bibitem{yeshwanthliu2023scannetpp}
Chandan Yeshwanth, Yueh-Cheng Liu, Matthias Nie{\ss}ner, and Angela Dai.
\newblock Scannet++: A high-fidelity dataset of 3d indoor scenes.
\newblock In {\em Proceedings of the International Conference on Computer
  Vision ({ICCV})}, 2023.

\bibitem{Yuan2022}
Yu-Jie Yuan, Yang-Tian Sun, Yu-Kun Lai, Yuewen Ma, Rongfei Jia, and Lin Gao.
\newblock Nerf-editing: Geometry editing of neural radiance fields, 2022.

\bibitem{Zhi2021}
Shuaifeng Zhi, Tristan Laidlow, Stefan Leutenegger, and Andrew~J. Davison.
\newblock In-place scene labelling and understanding with implicit scene
  representation, 2021.

\end{thebibliography}
\bibliographystyle{plain}
\end{document}